\documentclass[10pt,twocolumn,a4paper,fleqn]{article}
\usepackage[a4paper,left=15mm,right=15mm,top=30mm,bottom=25mm,columnsep=10mm]{geometry}
\usepackage{graphicx}
\usepackage{times}
\usepackage{titlesec}
\usepackage{indentfirst}
\usepackage[fleqn]{amsmath}
\usepackage{fancyhdr}
\usepackage{cite}
\usepackage{enumitem}
\usepackage{etoolbox}

\renewcommand{\normalsize}{\fontsize{10pt}{11.9pt}\selectfont} 
\titleformat{\section}
  {\normalfont\large\bfseries\MakeUppercase}{\thesection.}{0.5em}{}
\titlespacing*{\section}{0pt}{2.5ex plus 0.5ex minus .2ex}{1.5ex}

\titleformat{\subsection}
  {\normalfont\normalsize\bfseries}{\thesubsection.}{0.5em}{}
\titlespacing*{\subsection}{0pt}{1.5ex}{0.5ex}

\titleformat{\subsubsection}
  {\normalfont\normalsize}{\thesubsubsection.}{0.5em}{}
\titlespacing*{\subsubsection}{0pt}{2.0ex plus .5ex minus .2ex}{0.1ex}

\setlist{nosep}
\setlist[itemize]{leftmargin=*}
\setlist[enumerate]{leftmargin=*}
\setlist[description]{
  font=\itshape,
  leftmargin=0pt,
  labelsep=0.5em
}

\AtBeginEnvironment{table}{
  \scriptsize
}

\usepackage[small,bf]{caption}
\DeclareCaptionLabelSeparator{periodspace}{. }
\makeatletter
\newcommand{\affils}[1]{\def\@affils{#1}}
\renewcommand{\abstract}[1]{\def\@abstract{#1}}
\newcommand{\keywords}[1]{\def\@keywords{#1}}

\renewcommand{\@maketitle}{%
  \newpage
  \null
  \begin{center}
    {\fontsize{15pt}{15pt}\selectfont \bfseries \@title \par}
    \vskip 1.0em
    {\large \@author \par}
    \vskip 0.5em
    {\normalsize \@affils \par}
  \end{center}
  {\normalsize \noindent \textbf{Abstract:} \@abstract \par}
  \vskip 1em
  {\normalsize \noindent \textbf{Keywords:} \@keywords \par}
  \vskip 1.4em
}
\makeatother

\makeatletter
\renewenvironment{thebibliography}[1]{
  \section*{\refname}
  \normalsize
  \list{[\arabic{enumi}]}{
    \settowidth\labelwidth{[#1]}
    \leftmargin\labelwidth
    \advance\leftmargin\labelsep
    \setlength{\itemsep}{0pt}
    \setlength{\parsep}{0pt}
    \setlength{\topsep}{0pt}
    \setlength{\partopsep}{0pt}
    \usecounter{enumi}
    
  }
  \def\newblock{\hskip .11em plus .33em minus .07em}
  \sloppy\clubpenalty4000\widowpenalty4000
  \sfcode`\.=1000\relax
}{
  \endlist
}
\makeatother

\usepackage{bm}
\usepackage{subcaption}

\fancypagestyle{firstpage}{
  \fancyhf{}
  
  \fancyfoot[C]{\footnotesize \sffamily Author's version. In Proc. of the Joint Symposium of AROB 31st and ISBC 11th (AROB-ISBC 2026), pp.~923--927, 2026.}
}

\title{LLM-Guided Decentralized Exploration with Self-Organizing Robot Teams}

\author{Hiroaki Kawashima${}^{1}$, Shun Ikejima${}^{2}$, Takeshi Takai${}^{3}$, Mikita Miyaguchi${}^{3}$, and Yasuharu Kunii${}^{4}$}

\affils{${}^{1}$Graduate School of Information Science, University of Hyogo, Hyogo, Japan\\
(E-mail: kawashima@gsis.u-hyogo.ac.jp)\\
${}^{2}$School of Social Information Science, University of Hyogo, Hyogo, Japan\\
(E-mail: fa22a005.k@gmail.com)\\
${}^{3}$Takenaka Research \& Development Institute, Takenaka Corporation, Chiba, Japan\\
(E-mail: \{takai.takeshi, miyaguchi.mikita\}@takenaka.co.jp)\\
${}^{4}$Faculty of Science and Engineering, Chuo University, Tokyo, Japan\\
(E-mail: kunii@elect.chuo-u.ac.jp)\\
}
\abstract{%
When individual robots have limited sensing capabilities or insufficient fault tolerance, it becomes necessary for multiple robots to form teams during exploration, thereby increasing the collective observation range and reliability.
Traditionally, swarm formation has often been managed by a central controller; however, from the perspectives of robustness and flexibility, it is preferable for the swarm to operate autonomously even in the absence of centralized control. In addition, the determination of exploration targets for each team is crucial for efficient exploration in such multi-team exploration scenarios. 
This study therefore proposes an exploration method that combines (1) an algorithm for self-organization, enabling the autonomous and dynamic formation of multiple teams, and (2) an algorithm that allows each team to autonomously determine its next exploration target (destination). 
In particular, for (2), this study explores a novel strategy based on large language models (LLMs), while classical frontier-based methods and deep reinforcement learning approaches have been widely studied. The effectiveness of the proposed method was validated through simulations involving tens to hundreds of robots.
}


\begin{document}

\maketitle
\thispagestyle{firstpage}

\section{Introduction}

Exploring unknown environments such as lunar lava tubes carries various risks of robot failure due to communication disruptions, sensor malfunctions, and physical damage. In such scenarios, deploying a large number of small mobile robots, rather than relying on a few high-performance robots, can enhance the overall fault tolerance of the system.

To improve exploration efficiency, it is generally desirable for multiple robots to cover non-overlapping areas.
However, when individual robots have limited sensing capabilities or insufficient fault tolerance, it becomes necessary for multiple robots to form teams during exploration so as to increase their collective observation range and reliability. Traditionally, swarm formation has often been managed by a central controller; yet, from the perspectives of robustness and flexibility, it is preferable for the swarm to operate autonomously even in the absence of centralized coordination.

In this study, we propose an exploration framework that integrates (1) a self-organization algorithm that autonomously and dynamically forms multiple teams, and (2) a destination-selection algorithm that enables each team to decide its exploration target. Once teams are formed, each team must determine its next exploration destination. While classical frontier-based methods and deep reinforcement learning approaches have been widely studied, we explore a novel strategy based on large language models (LLMs). 

Specifically, each team leader selects a target cell from the set of frontier cells using information extracted from the probabilistic occupancy grid map, the current position of the team, and the positions and destinations of other teams. Using this information, the LLM performs common-sense reasoning to determine the team's next destination. Our contribution lies in demonstrating in simulation the effectiveness of LLMs in decentralized multi-robot exploration, particularly in scenarios involving dynamic team formation.

\section{Related work}

Multi-robot exploration of unknown environments is a long-standing research topic and various techniques have been proposed, including classic frontier-based methods~\cite{yamauchiAGENTS1998}, randomized graph search~\cite{franchiICRA2007,orioloICRA2004}, optimization with utility~\cite{burgard2005TRO}. 
Recently, deep reinforcement learning methods have also been applied to multi-robot exploration~\cite{li2020TNNLS}, and extended to asynchronous and decentralized settings~\cite{yuAsyncRL2023}. While these methods focus on individual robot behavior, our approach emphasizes team formation and destination selection in a decentralized manner. In addition, we also explore the use of LLMs for destination selection, which is a novel approach in this context.

\section{Decentralized exploration}
\subsection{Robot model and environment representation}

\begin{figure}[t]
    \centering
    \begin{subfigure}[t]{0.50\linewidth}
        \centering
        \includegraphics[width=\linewidth]{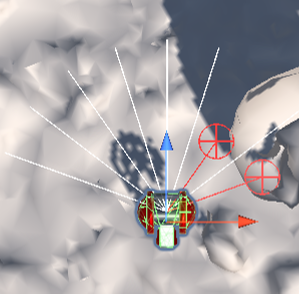} 
        \caption{Ray sensor}
        \label{fig:sensor}
    \end{subfigure}
    \begin{subfigure}[t]{\linewidth}
        \centering
        \includegraphics[width=\linewidth]{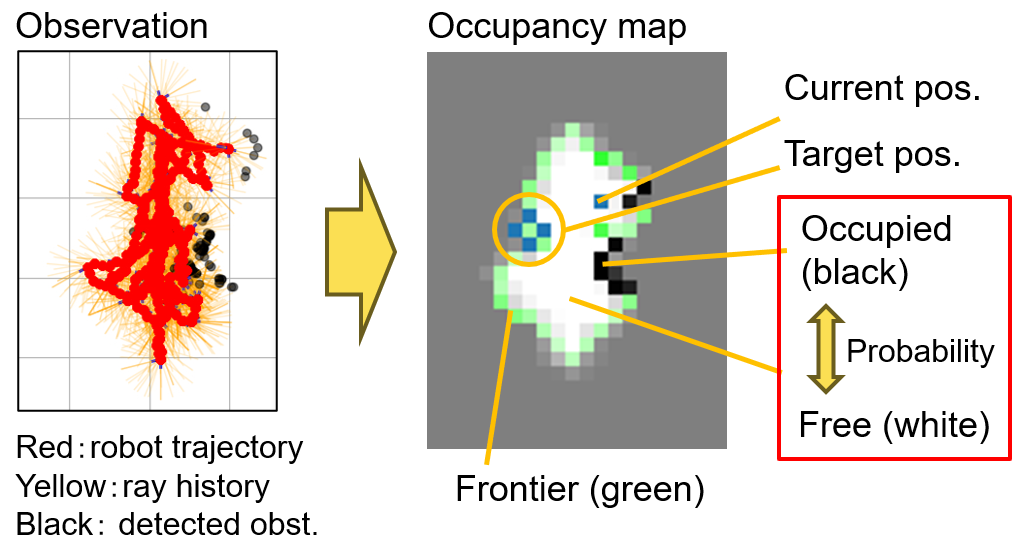} 
        \caption{Gridmap generation}
        \label{fig:gridmap}
    \end{subfigure}
    \caption{~Sensor model and occupancy grid map generation. In (b), occupancy probability values are indicated in grayscale: gray cells represent unexplored areas, white cells represent explored free areas (free space), and black cells represent explored areas with obstacles (occupied space). Green cells represent frontier cells.}
    \label{fig:sensor_and_gridmap}
\end{figure}

This study addresses the problem of exploring an unknown environment using a swarm of $N$ mobile robots equipped with short-range, sparsely sampled sensors. All experiments are conducted in simulation. 
Each robot is approximately 0.3~m in diameter. To simulate limited sensing capability, each robot is provided with nine rays for detecting obstacles and free space in its vicinity, as illustrated in Fig.~\ref{fig:sensor}. The sensing fan spans 70 degrees to both the left and right from the forward direction, and the maximum sensing range of each ray is 1~m.

Each robot constructs a probabilistic occupancy grid map based on its sensor observations, as illustrated in Fig.~\ref{fig:gridmap}. The grid map is composed of 0.5~m $\times$ 0.5~m cells, and each cell stores the probability of being occupied by an obstacle. The robots update these occupancy probabilities using Bayesian filtering, implemented as additive updates in the log-odds representation based on their range measurements, which are subject to higher uncertainty, and their own positions, which are assumed to be more reliable. 

Because communication-range limitations are also an important factor in real-world scenarios, each robot continuously maintains its own local map and, when it encounters nearby robots, integrates the received maps with its own using additive updates in the log-odds representation.
In this study, however, we assume full communication among all robots so that we can concentrate on our primary objectives: team formation and destination selection. Under this assumption, all local maps are fused into a single global map at each step, and this global map is shared among every robot.
We also assume that each robot is aware of the positions of their own and other robots for simplicity, although in real-world scenarios, relative positions would need to be estimated using onboard sensors and communication.

\subsection{Task description and team behavior}

The objective of the exploration is to maximize the area covered within a fixed time frame. Initially, the robots have no prior knowledge of the environment. Frontier cells, defined as the boundary between explored free space and unexplored space, are therefore identified as potential exploration targets, and one of these frontier cells is selected as the next exploration target. Here, thresholds are applied to the occupancy probabilities to classify cells as free, occupied, or unknown.
Each robot computes a route to its assigned target using the A$^\ast$ algorithm based on the current occupancy grid map, where the probability values are used as the cost. Although the map is represented by grid cells, the robots move in continuous space at each step using discrete actions: stop, fixed forward motion, and fixed left/right rotations. The forward motion at a step is less than the cell size (0.2 m/step) and the rotations are 40 degrees/step.

We introduce teams of multiple robots during exploration to increase collective observation range and reliability. These teams are formed dynamically in a decentralized manner, and each team autonomously determines its exploration target. Within each team, a leader robot is designated to select a target cell on behalf of the team, and this target is shared among all team members. Each member then independently plans its route to the shared target and moves accordingly. After any member reaches the assigned target, the leader selects a new target and the process is repeated.

We also assume that each robot has a limited battery capacity and operates in one of two modes: \textit{Explore (EXP)} or \textit{Charge (CHR)}.
When the battery level falls below a specified threshold, the robot switches to CHR mode, leaves its current team, and heads to a charging station to recharge. After completing the recharge, the robot switches back to EXP mode and resumes exploration by autonomously forming a new team or joining an existing one.

The following subsections describe these two components in detail: decentralized team formation and LLM-based team destination determination.

\subsection{Decentralized team formation}

The optimal team formation depends on various parameters, such as the tasks to be performed (e.g., exploration, transport objects), operational conditions, and the robots' states (e.g., position, relative configuration, battery charge level). Therefore, we propose an autonomous formation algorithm from the perspective of individual robots' behavioral policies.

To enable decentralized team formation, we introduce a concept of \textit{desired team size} $\tilde{n}_i$ as an internal state parameter of each robot $i$. This allows each robot to autonomously determine the desirable number of robots to form a team based on the current situation. For instance, when moving into unexplored areas, a larger desired team size is preferable because a certain number of robots is required. In this situation, where the desired size exceeds the actual team size $n_i$, i.e., $\tilde{n}_i > n_i$, the robot enters a recruitment state. It attempts to join with other individuals or teams that are also in this recruitment state and meet certain conditions, such as being nearby.
Conversely, when smaller teams are advantageous, the robot decreases the desired size. For example, when leaving the team to return to a charging station, it is preferable to act alone, and the robot sets the desired team size to 1. If the desired size is smaller than the actual team size, i.e., $\tilde{n}_i < n_i$, the robot attempts to leave the team. The conditions for leaving depend on scenarios, and we describe a simple implementation example in Sec.~\ref{sec:exp-selforganization}.

\subsection{LLM-based team destination determination}

Regarding destination determination, while classical frontier-based methods and deep reinforcement learning approaches exist, this study explores a novel approach using large language models (LLMs). 

The LLM, a pre-trained model without fine-tuning, is provided with a cell coordinate list that includes the occupancy status of the following information:
\begin{enumerate}
    \item explored non-obstacle cells
    \item explored obstacle cells
    \item frontier cells
\end{enumerate}
as well as the current position of the team leader and the positions and destinations of other teams. 
The positions of team leader robots are used as the current position of each team, while any other definition of team position (e.g., team center) is also possible.

Each cell in the list includes the distance from the team leader's current position and neighborhood features: the information for each cell consists of (coordinate, label, neighborhood features, distance).
Here, the label indicates whether the cell is an explored non-obstacle cell (1), an explored obstacle cell (2), or a frontier cell (3).
For the neighborhood features, we adopted the number of frontier cells and occupied cells within 8-connected neighboring cells.

Using this information with a given prompt template, the LLM performs common-sense reasoning to determine the team's next destination from frontier cells. Specifically, at each decision step, the team leader constructs a prompt for the LLM that includes the current map information (i.e., a cell list explained above), the team's current position, and the positions and destinations of other teams, and requests the LLM to decide a next target frontier cell for ``efficient exploration.''

\section{Experiments}

\subsection{Setup}

The exploration environment was modeled after future lunar lava tubes, using 3D mesh data obtained from terrestrial lava tube structures on the Earth. For LLM, Azure OpenAI gpt-4o was used. 
If the returned cell is not a frontier cell, the API was called again until a valid frontier cell was obtained.
The maximum number of retries for API calls was set to 5, after which the baseline method below was used as a fallback.

\noindent{\textbf{Baseline target selection:}} As a baseline for destination selection, we implemented a probabilistic sampling method that selects a frontier cell nearby each team, incorporating individual biases assigned to each robot; consequently, destination selections are influenced by the specific characteristics of the current team leaders.
Specifically, the distance from the leader's position to each frontier cell is calculated first. A quantile value is then sampled from a normal distribution with a robot-specific biased mean and a standard deviation of $0.02$, clipped to the range $[0, 1]$. For each robot, this mean is uniformly sampled from the range $[0.10, 0.25]$ at the start of each simulation.
The frontier cell at the corresponding distance quantile is selected as the target. This approach allows for the occasional selection of more distant frontiers rather than always choosing the nearest one, thereby promoting exploration diversity.

\subsection{Self-organization of teams}
\label{sec:exp-selforganization}

To verify the basic concept of the decentralized team formation, we implemented a simple set of the desired team size based on the robots' two operational modes: EXP and CHR. In EXP mode, the desired team size was set to 5, whereas in the CHR mode, it was set to 1.
In the experiments, we specifically applied the following simple rules.
Merging is performed at the team level, whereas splitting is performed at the individual robot level.
Merging takes place when two teams are both in the recruitment state and within a certain proximity. Specifically, if robot $i$ finds robot $j \notin T(i)$ within a threshold distance $d_{\text{join}}$, and the condition $n_i + n_j \leq \min(\tilde{n}_i, \langle\tilde{n}_j\rangle_{j \in T(j)})$ is satisfied, then teams $T(i)$ and $T(j)$ are merged. For this process, single robots are treated as teams of size one.
Conversely, leaving is performed by individual robots, and is permitted unconditionally when the battery charge level is low (i.e., in CHR mode). For more general situations, a leaving rule could be designed such that a robot is allowed to depart if the average desired team size within the team is smaller than the actual team size: Robot $i$ could leave when $\langle\tilde{n}_i\rangle_{i \in T(i)} < n_i$ holds.

\begin{figure}[t]
    \centering
    \includegraphics[width=\linewidth]{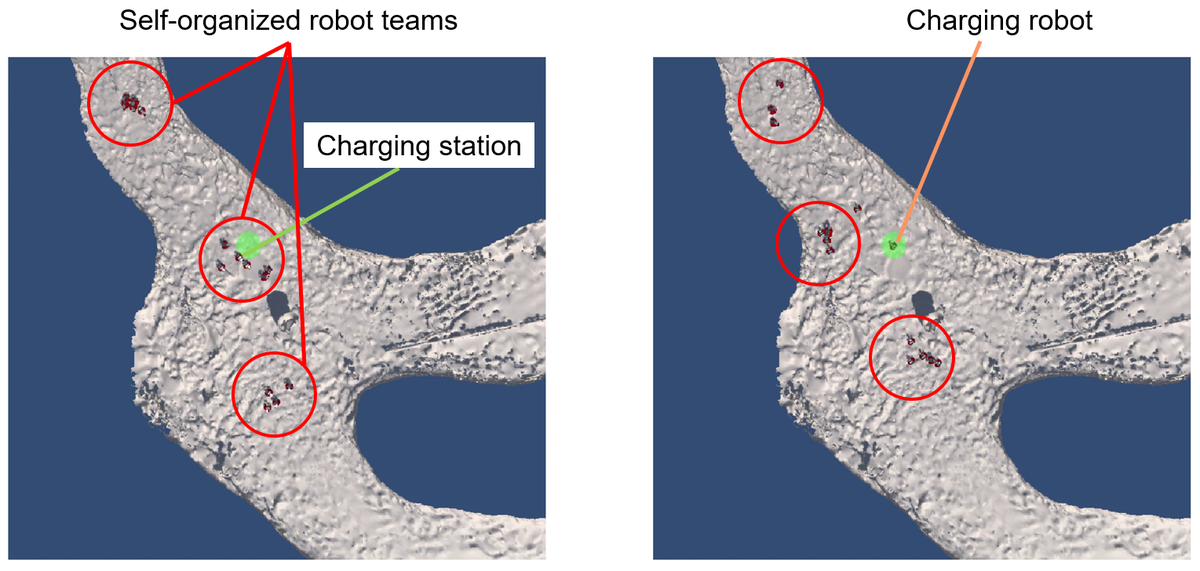} 
    \caption{Example of self-organized teams and a charging robot}
    \label{fig:selforganizing_example}
\end{figure}

Before evaluating the LLM-based destination selection method, we first qualitatively verify the effectiveness of exploration using the baseline method integrated with the autonomous team formation algorithm.
Figure~\ref{fig:selforganizing_example} illustrates an example where $N = 15$ robots dynamically manage the two tasks of team organization and battery recharging.
As shown in this figure, robots in EXP mode autonomously form teams (indicated by red circles), whereas a robot in CHR mode leaves its team and proceeds to the charging station (indicated in green) to recharge, as shown in Fig.~\ref{fig:selforganizing_example} (right).

\begin{figure}[t]
    \centering
    \includegraphics[width=\linewidth]{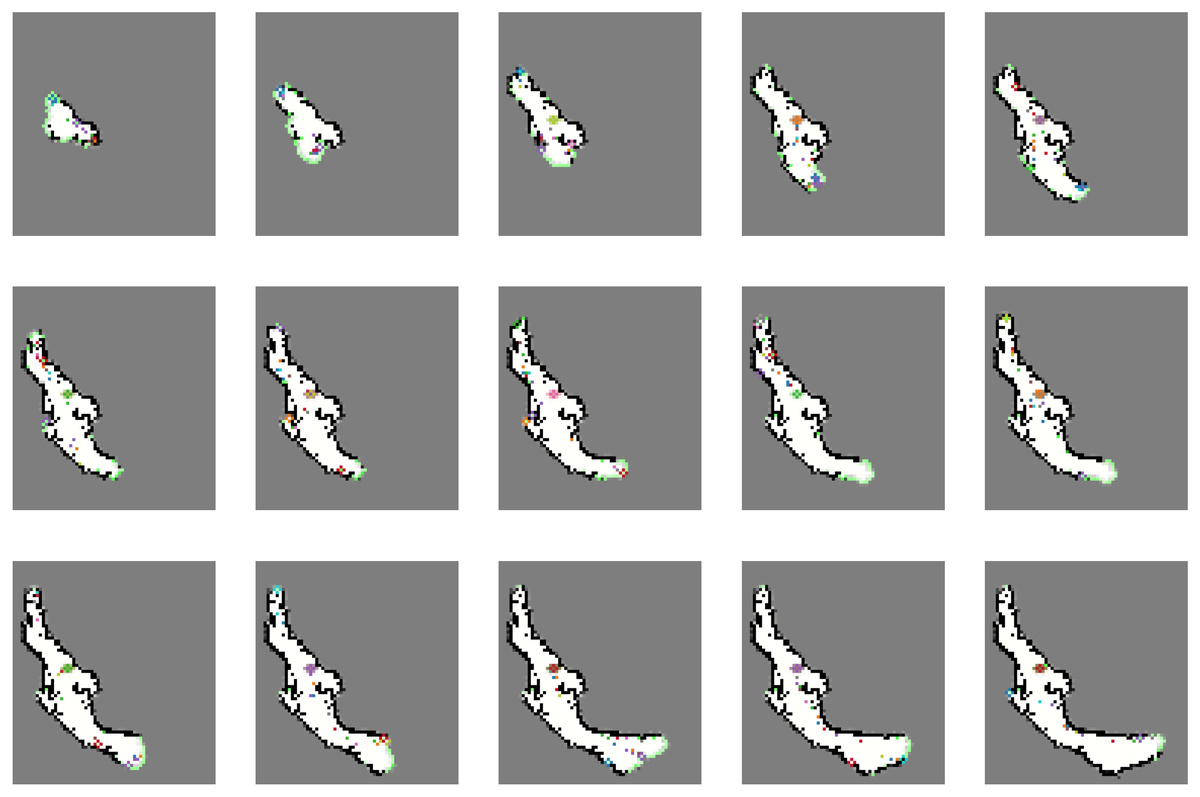}
    \caption{~Example exploration using the baseline method}
    \label{fig:baseline_exploration}
\end{figure}

Figure~\ref{fig:baseline_exploration} illustrates an example of how the explored area expands over time using the baseline method with the same condition above. The figure clearly shows that explored free area (white) progressively increases, while unexplored area (gray) decreases as the exploration proceeds. Throughout the process, multiple teams are dynamically formed and reorganized as robots join or leave teams, eventually returning to the charging station individually.

\subsection{LLM-based destination selection}

\begin{figure}[t]
    \centering
    \includegraphics[width=\linewidth]{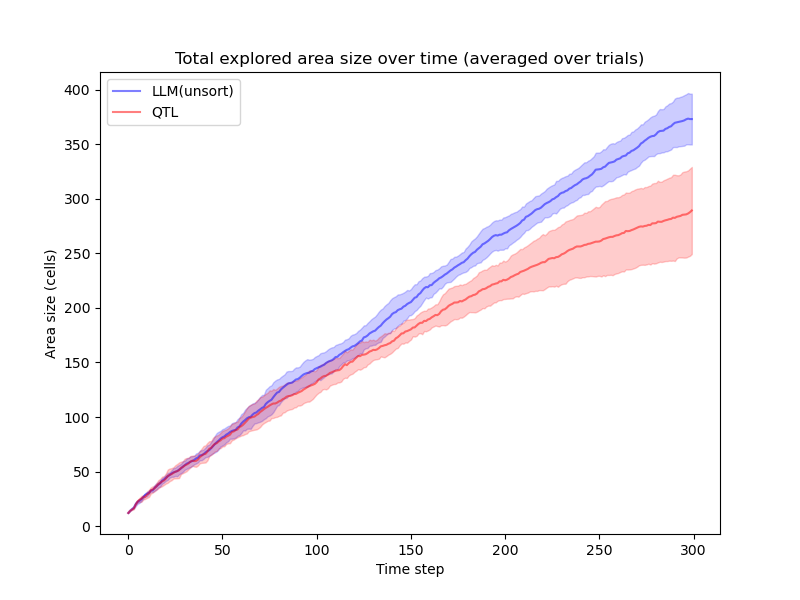} 
    \caption{~Comparison of explored area between baseline and LLM-based methods (N = 15)}
    \label{fig:llm_exploration_comparison}
\end{figure}

To quantitatively evaluate the effectiveness of the proposed LLM-based destination selection method, we compared the total explored area achieved within a fixed duration of 300 steps using $N = 15$ robots. The LLM-based method was initiated after 20 steps of baseline exploration; this ``warm-up period'' ensures that sufficient map information was available for the LLM to make informed decisions. The results, averaged over five independent trials, are presented in Fig.~\ref{fig:llm_exploration_comparison}. As illustrated in the figure, the proposed LLM-based method achieved approximately a 20\% increase in the explored area compared to the baseline.

\begin{figure}[t]
    \centering
    \includegraphics[width=\linewidth]{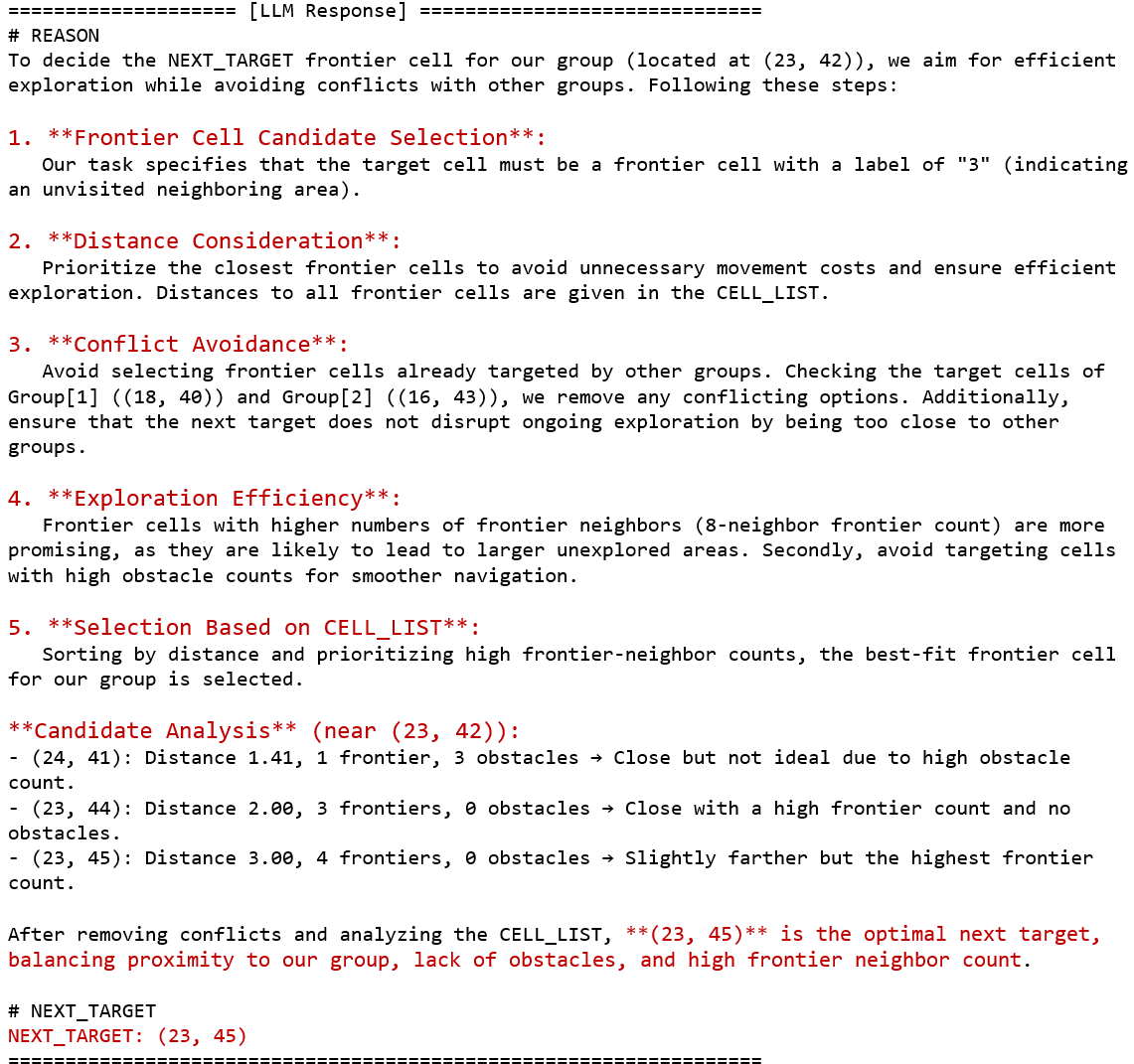} 
    \caption{~Example of LLM reasoning for destination selection}
    \label{fig:llm_reasoning}
    \vspace{5mm}
\end{figure}

An example of LLM reasoning is shown in Fig.~\ref{fig:llm_reasoning}. 
As shown in the figure, the LLM successfully incorporates multiple factors, including proximity to its own team, avoidance of overlap with other teams' target cells, and the presence of nearby frontier (a larger number of frontier cells is preferable) and obstacle cells (a smaller number of obstacle cells is preferable), to generate a set of candidate target cells. After summarizing the features of these candidates, the LLM selects the most suitable target cell, which is not necessarily the closest frontier cell. This example demonstrates the LLM's ability to perform common-sense reasoning based on the provided occupancy grid map and team-level information.

\begin{figure}[t]
    \centering
    \includegraphics[width=\linewidth]{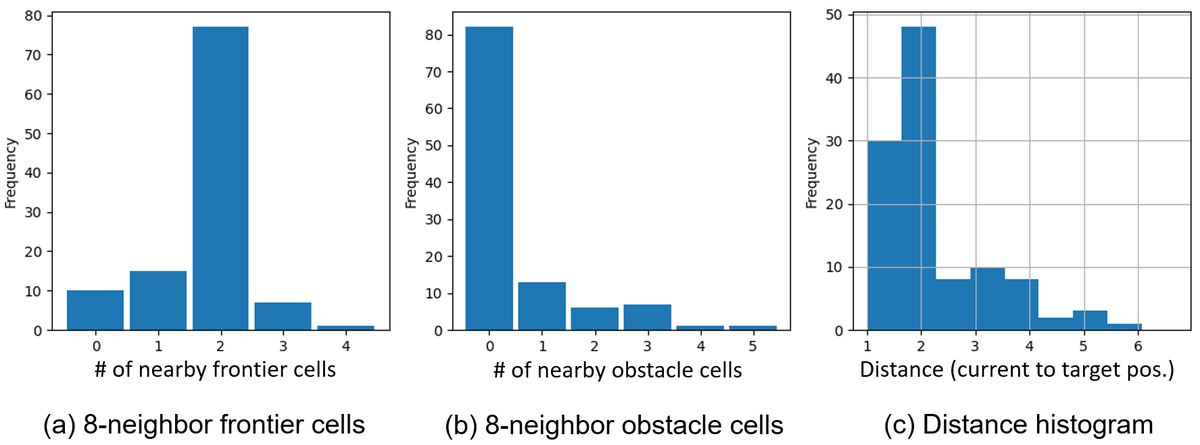} 
    \caption{~Comparison of target cell statistics between baseline and LLM-based methods}
    \label{fig:llm_targetstats}
\end{figure}

Figure~\ref{fig:llm_targetstats} shows the distribution of nearby frontier cells (Fig.~\ref{fig:llm_targetstats}a), the distribution of nearby obstacle cells (Fig.~\ref{fig:llm_targetstats}b), and the histogram of distances from the team leader to the target cell (Fig.~\ref{fig:llm_targetstats}c). These distributions clearly illustrate the characteristics of the target cells selected by the LLM: a preference for target cells surrounded by more frontier cells and fewer obstacle cells, while allowing slightly more distant target cells.

\subsection{Scalability with larger numbers of robots}

\begin{figure}[t]
    \centering
    \includegraphics[width=\linewidth]{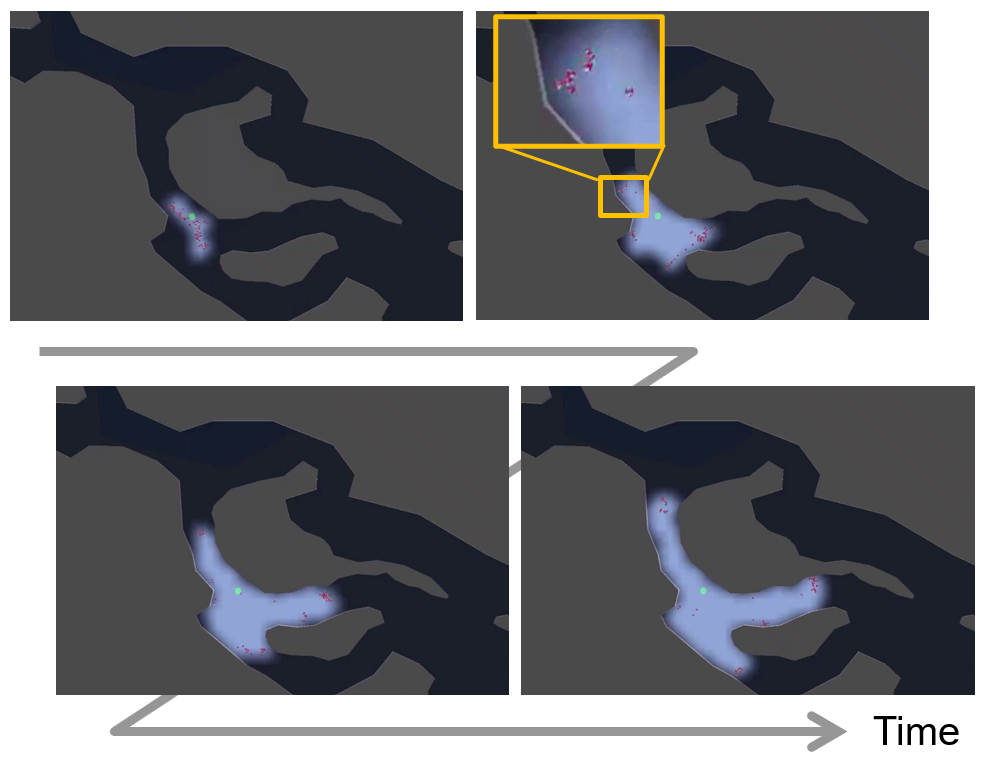} 
    \caption{~Exploration with autonomous team formation with $N=50$. The bright areas represent explored regions and the small red dots correspond to individual robots.}
    \label{fig:base_exploration_N50}
    \vspace{5mm}
\end{figure}

We validated the proposed method with larger numbers of robots, combined with the autonomous team formation mechanism described in Sec.~\ref{sec:exp-selforganization}.
With $N=50$ robots, we confirmed that the baseline exploration with the autonomous team formation performs effectively, enabling exploration over a wide area (Fig.~\ref{fig:base_exploration_N50}). The black areas represent lava tubes, while the bright areas indicate regions explored by the robots. A number of small red dots correspond to individual robots operating within the 3D simulator, each autonomously forming teams within the robot group while advancing the exploration.

\begin{figure}[t]
    \centering
    \includegraphics[width=\linewidth]{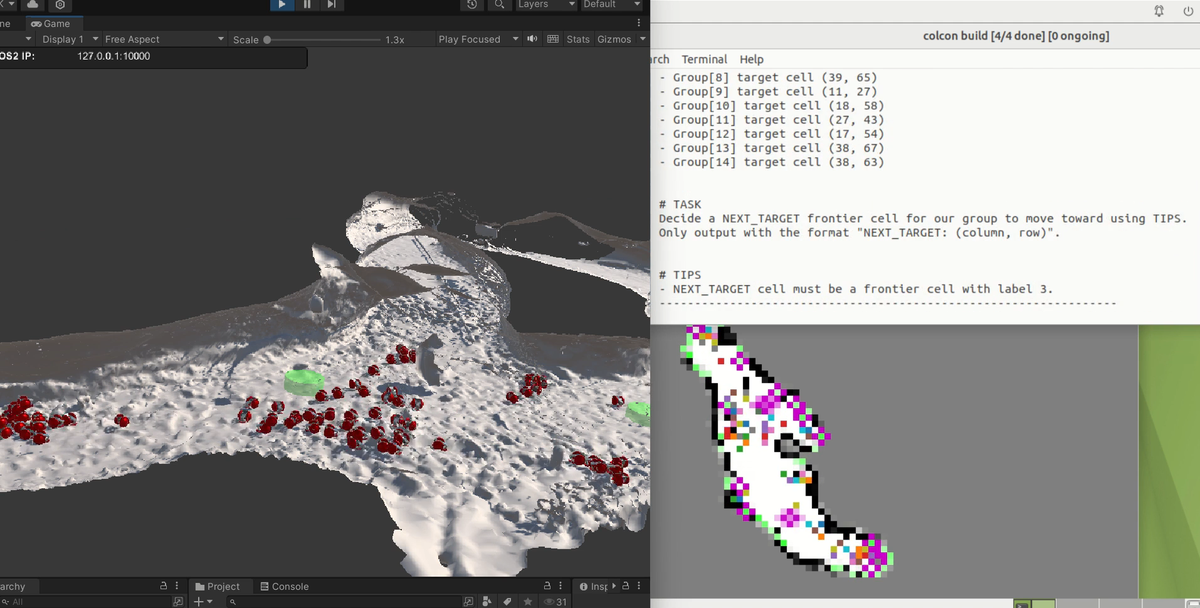} 
    \caption{~A screenshot of the LLM-based exploration simulation with $N=100$}
    \label{fig:llm_exploration_N100_screenshot}
\end{figure}

Furthermore, we validated the scalability of the proposed LLM-based exploration with self-organizing team formation by increasing the number of robots to $N=100$. Figure~\ref{fig:llm_exploration_N100_screenshot} presents a screenshot of the simulation. The left window displays the 3D simulator, in which numerous small red objects represent individual robots exploring the lava tube environment.
The lower-right window shows the occupancy map obtained through exploration, with a color scheme consistent with Fig.~\ref{fig:sensor_and_gridmap}. In this map, cells surrounded by purple dots indicate the target cell of the respective teams, while dots in other colors represent the positions of individual robots. 

\section{Conclusion}

This paper presents a decentralized exploration method that integrates (1) an autonomous team formation algorithm and (2) an LLM-based destination selection algorithm. The effectiveness of the proposed method was validated through simulations by comparing it with a baseline exploration method utilizing probabilistic frontier sampling. Furthermore, we qualitatively confirmed that large-scale robot swarms equipped with team-level LLM reasoning can efficiently expand the explored area. Future work includes experiments under limited communication conditions, policy learning to adapt the team sizes according to environmental conditions and robots' internal states, and decentralized task switching to other objectives, such as object transport.

\section*{Acknowledgment}

This work was supported by JST Moonshot R\&D Program Grant Number JPMJMS2238, Japan.

\end{document}